\documentclass[10pt,twocolumn,letterpaper]{article}

\usepackage{cvpr}              % To produce the CAMERA-READY version
\usepackage[dvipsnames]{xcolor}

\definecolor{cvprblue}{rgb}{0.21,0.49,0.74}
\usepackage[pagebackref,breaklinks,colorlinks,citecolor=cvprblue]{hyperref}
\usepackage{multirow}    % tables multirow
\usepackage{xcolor}
\usepackage{comment}

\usepackage{multicol} % Add this to your document preamble

\definecolor{lightred1}{RGB}{255, 204, 204} % A very light red
\definecolor{lightred2}{RGB}{255, 153, 153} % A light-medium red
\definecolor{lightred3}{RGB}{255, 102, 102} % A medium-light red
\definecolor{MyLightGray}{rgb}{0.95, 0.95, 0.95}
\definecolor{MyLightGreen}{rgb}{0.87, 0.93, 0.917}
\definecolor{MyLightPink}{rgb}{0.957, 0.87, 0.914}

\definecolor{ResNetBlue}{rgb}{0.192, 0.454, 0.643}
\definecolor{kNNOrange}{rgb}{0.89, 0.50, 0.169} % rgba(227,128,43,255)
\definecolor{EMDRed}{rgb}{0.757, 0.243, 0.239} % rgba(193,62,61,255)
\definecolor{CHMBrown}{rgb}{0.517, 0.36, 0.322} % rgba(132,92,82,255)

\newcommand{\subsec}[1]{\noindent\textbf{#1}~~}
\newcommand{\class}[1]{{\footnotesize\textsf{#1}}\xspace}
\definecolor{customdarkred}{RGB}{200,0,0}

\def\paperID{14} % *** Enter the Paper ID here
\def\confName{CVPR}
\def\confYear{2024}

\title{Allowing humans to interactively guide machines where to look does \emph{not}\\always improve human-AI team's classification accuracy}

\author{Giang Nguyen\textsuperscript{1, *}, Mohammad Reza Taesiri\textsuperscript{2, *}, Sunnie S. Y. Kim\textsuperscript{3}, Anh Totti Nguyen\textsuperscript{1}\\
\textsuperscript{1}Auburn University, \textsuperscript{2}University of Alberta, \textsuperscript{3}Princeton University\\
{\tt\small \textsuperscript{1}nguyengiangbkhn@gmail.com, \textsuperscript{2}mtaesiri@gmail.com, \textsuperscript{3}suhk@princeton.edu, \textsuperscript{1}anh.ng8@gmail.com}\\
\small \textsuperscript{*}Equal contributions
}

\begin{document}
\maketitle
\begin{abstract}
Via thousands of papers in Explainable AI (XAI), attention maps \cite{vaswani2017attention} and feature importance maps \cite{bansal2020sam} have been established as a common means for finding how important each input feature is to an AI's decisions.
It is an interesting, unexplored question whether allowing users to edit the feature importance at test time would improve a human-AI team's accuracy on downstream tasks.
In this paper, we address this question by leveraging CHM-Corr, a state-of-the-art, ante-hoc explainable classifier \cite{taesiri2022visual} that first predicts patch-wise correspondences between the input and training-set images, and then bases on them to make classification decisions.
We build CHM-Corr++, an interactive interface for CHM-Corr, enabling users to edit the feature importance map provided by CHM-Corr and observe updated model decisions.
Via CHM-Corr++, users can gain insights into if, when, and how the model changes its outputs, improving their understanding beyond static explanations.
However, our study with 18 expert users who performed 1,400 decisions finds no statistical significance that our interactive approach improves user accuracy on CUB-200 bird image classification over static explanations. 
This challenges the hypothesis that interactivity can boost human-AI team accuracy~\cite{sokol2020one,sun2022exploring,shen2024towards,singh2024rethinking,mindlin2024beyond,lakkaraju2022rethinking,cheng2019explaining,liu2021understanding} and raises needs for future research. 
We open-source CHM-Corr++, an interactive tool for editing image classifier attention (see an interactive demo \href{http://137.184.82.109:7080/}{here}).
% , and it lays the groundwork for future research to enable effective human-AI interaction in computer vision. 
We release code and data on \href{https://github.com/anguyen8/chm-corr-interactive}{github}.

\end{abstract}    
\section{Introduction}
\label{sec:intro}

\begin{figure}[t]
    \centering
    \includegraphics[width=1.0\linewidth]{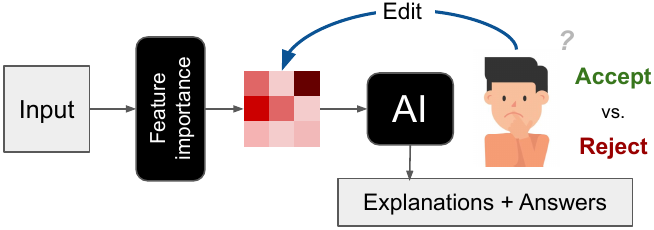}
    \caption{Users provide \textcolor{blue!60!black}{\textbf{feedback}} to the explainable AI by editing the feature importance map (like an ``attention map'' \cite{vaswani2017attention}).}
    \label{fig:teaser}
\end{figure}

% Understanding the inner workings of Deep Neural Networks (DNNs) has long been considered the holy grail in the field of Artificial Intelligence (AI).
Despite much attention from the community, 
% Despite significant efforts in developing tools for DNNs' interpretability—ranging from methods that ensure models are interpretable by design to post-hoc explanations that elucidate model decisions after the fact—the limited 
the practical utility of Explainable AI (XAI) tools in downstream applications (\eg image classification~\cite{ford2022explaining,kim2022hive,nguyen2021effectiveness,shitole2021one}) remains limited, hindering human-AI collaboration in real-world settings.
\begin{comment}
Techniques such as feature attribution maps, which have been proposed and harnessed in thousands of XAI papers, highlight the specific input features that a model relies on for its predictions but provide merely a static snapshot of the model's thought process at a given moment.
Similarly, example-based explanations, such as nearest neighbors, also present a static view, encapsulating the model's reasoning within a fixed set of samples.
Yet, we have recently witnessed the emergence of effective XAI approaches, notably through correspondence-based explanations—a combination of attribution maps and exemplars, which have shown promise in enhancing the performance of human-AI teams beyond what either AI or humans could achieve independently~\cite{taesiri2022visual}.
% Explaining via examples~\cite{kenny2022towards, kenny2023advancing,liu2022learning,nguyen2021effectiveness,taesiri2022visual,nguyen2023advisingnets} 
This type of explanations offers two main benefits: 1) It provides additional contextual information beyond the mere input image (as found in feature attribution maps~\cite{selvaraju2017grad}) via support samples from an external knowledge base, such as model training set~\cite{nguyen2021effectiveness,taesiri2022visual,nguyen2023advisingnets}.
2) It facilitates a compare-and-contrast approach for human users by pinpointing the patches the model focuses on in both the input and support samples to determine the image label (see Fig.~\ref{fig:chm_corr++}--{\setlength{\fboxsep}{0pt}\colorbox{MyLightGreen}{\strut\textbf{b}}}).
\end{comment}
A major \textbf{limitation} is that there is no interface for humans to provide feedback to the model so that it can update its decisions, which could change users' thoughts and final decisions.
For example, feature importance maps \cite{bansal2020sam,chen2022gscorecam} and example-based explanations \cite{nguyen2021effectiveness} are among the most popular XAI methods that offer insights into ``what a model is looking at'' and which real examples support a model decision, respectively.
However, they only offer a \textbf{static, one-time} explanation of the input.
Here, we test allowing users to edit the ``attention'' input to the model and observe updated model decisions iteratively until they are ready to make the final decisions (\cref{fig:teaser}).

We perform our study on CHM-Corr, a recent explainable CUB-200 bird classifier \cite{taesiri2022visual} that combines the best of both worlds by first finding, at the patch level, how the input image is similar to the nearest training-set examples, and then using these patch-wise correspondences to predict the image label (see Fig.~\ref{fig:chm_corr++}--{\setlength{\fboxsep}{0pt}\colorbox{MyLightGreen}{\strut\textbf{b}}} and Supp.~\ref{sec:Qualitative_figures} for examples of CHM-Corr explanations).
CHM-Corr explanations enabled users to achieve \textbf{state-of-the-art human-AI team accuracy} in bird identification on CUB-200 \cite{taesiri2022visual}.

We build an interactive interface called CHM-Corr++ for CHM-Corr, allowing users to manipulate the ``attention'' of the CHM-Corr classifier by selecting the set of patches that the classifier uses in its decision-making step (\cref{fig:teaser}).
By iteratively telling the model where to look, and observing the changes in the output space (see Fig.~\ref{fig:chm_corr++}), users could better understand the AI model and make more informed and accurate decisions.
% human users could understand the AI more and make decisions more correctly.
% Surprisingly, in a user study involving 1400 decisions with machine learning (ML) experts, we find that our interactive interface did not help participants perform better than when using static, one-off correspondence explanations for binary decision-making tasks (i.e., accepting or rejecting the AI model's top-1 label) in CUB-200 image classification.
% Following this result, we investigate the reasons for the ineffectiveness of dynamic explanations and highlight future research avenues to improve their utility.
% We summarize the main contributions of this work as follows:

Via a user study of 1400 decisions, surprisingly, we did not find the interactivity to help improve users' decision-making accuracy (\cref{tab:main_table}). 
This finding is intriguing and in stark contrast to the common hypothesis that interactive explanations might improve human-AI collaboration effectiveness and therefore human-AI team accuracy \cite{shen2024towards,singh2024rethinking,mindlin2024beyond,lakkaraju2022rethinking}.

% focus on dynamic explanations, emphasizing the evolving nature of explanations as they adapt over time to the interactions between human users and an AI model (see Fig.~\ref{fig:chm_corr++}). 
% In computer vision, the dynamics of visual explanations have been relatively underexplored.
% Zhang \etal~\cite{zhang2023may} allow users to interact with AI models to select the most accurate classifiers from a set of options, often through dialogues and feature attribution~\cite{zhang2023may}.
% In comparison, our work tasks humans with evaluating the accuracy of labels predicted by an image classifier.
% Ours is also orthogonal with previous interactive XAI works
% on tabular and textual data~
% \cite{hohman2019gamut,cheng2019explaining,tenney2020language}, where users can modify input features to see how these changes affect the model's output alongside existing feature attributions. 
% Our method does not involve altering the input image directly; instead, it allows users to guide the AI model's focus towards particular features of objects that are deemed informative for classification, thereby steering the model's attention and influencing its decision-making process. 

\begin{figure*}[!hbt]
    \centering
    \includegraphics[width=0.98\linewidth]{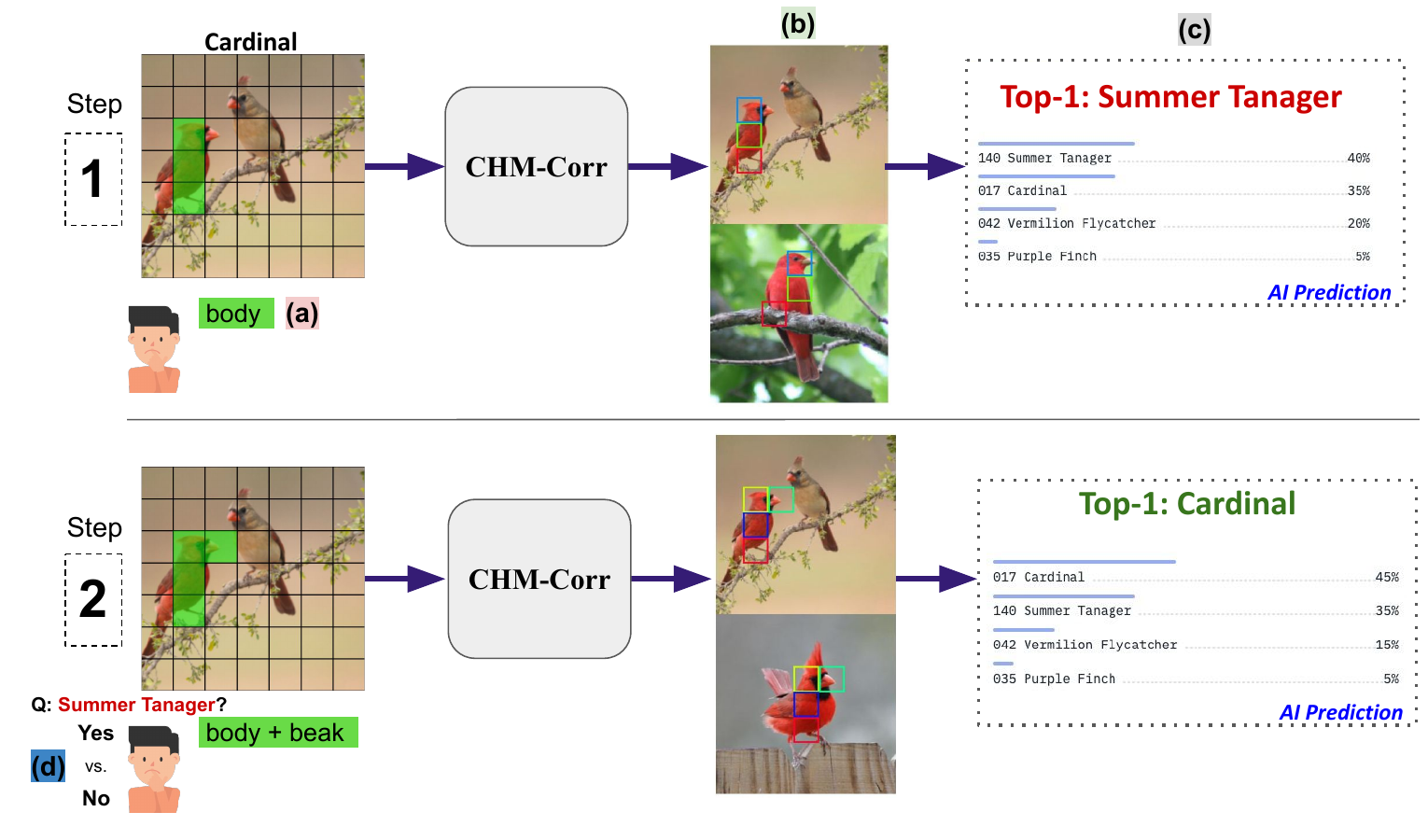}

    \caption{Our CHM-Corr++ interactive interface. 
    We let users interact with the image classification model (here CHM-Corr~\cite{taesiri2022visual}) via controlling the attention (selecting patches) the model should look at {\setlength{\fboxsep}{0pt}\colorbox{MyLightPink}{\strut(\textbf{a})}}. 
    Based on the user-guided attention, the model compares the input image (GT class: \class{Cardinal}) with candidate, training examples to simultaneously generate visual-correspondence explanations {\setlength{\fboxsep}{0pt}\colorbox{MyLightGreen}{\strut(\textbf{b})}} and predictions {\setlength{\fboxsep}{0pt}\colorbox{gray!15}{\strut(\textbf{c})}}. 
    The user \textbf{iteratively} observes the dynamic explanations {\setlength{\fboxsep}{0pt}\colorbox{MyLightGreen}{\strut(\textbf{b})}} and predictions {\setlength{\fboxsep}{0pt}\colorbox{gray!15}{\strut(\textbf{c})}} to understand the image classification model to accept or reject {\setlength{\fboxsep}{0pt}\colorbox{ResNetBlue}{\strut(\textbf{d})}} the \emph{original} top-1 predicted label (here \textcolor{customdarkred}{\textbf{Summer Tanager}}).}

    \label{fig:chm_corr++}
\end{figure*}

% \begin{itemize}
    % \item We develop and open-source an interactive interface called CHM-Corr++ (\cref{fig:chm_corr++}) that allows users to directly interact with an image classification model (CHM-Corr~\cite{taesiri2022visual}) by controlling model attention. 
    % At a high level, our interface presents model attention, support samples, and their respective classification outputs.
    
    % \item Via a user study of 1400 decisions, surprisingly, we did not find the interactivity to help improve users' decision-making accuracy (\cref{tab:main_table}). 
    % This finding is unexpected yet intriguing and valuable for the community because it questions the prevailing assumption that interactive medium represents the next frontier in XAI~\cite{shen2024towards,singh2024rethinking,mindlin2024beyond,lakkaraju2022rethinking}.
    
    % \item We provide an in-depth analysis and insights into how users interact with explanations, both static and dynamic, 
    % and identify settings where dynamic explanations improve vs. hurt users' decision-making (\cref{table:accuracy_conditions}).
    % and concur that despite being intuitive, dynamic explanations seriously hurt human decision-making in two specific scenarios (ii) and (iii) in Table~\ref{table:accuracy_conditions}, making their users unable to make more accurate decisions in general. 
    % Our finding paves the way for future research in creating better interactive human-AI mediums.
    
% \end{itemize}
\section{Related Work}
\label{sec:relatedworks}

% {\color{red}{Could be substantially shortened for a 4-page submission.}}

Fine-grained visual classification is a domain with active XAI research.
Numerous explanations methods have been proposed, producing explanations of various forms. 
Representative forms include heatmaps~\cite{li2021scouter,Pillai_2022_CVPR}, examples~\cite{goyal2019ICML,vandenhende2022making,taesiri2022visual,nguyen2023advisingnets}, concepts~\cite{ramaswamy2022elude,koh2020concept,yuksekgonul2023posthoc}, and prototypes~\cite{chen2019protopnet,donnelly2022deformable,nauta2021prototree}.
Regardless of the form, however, most explanations are \emph{static}.
They are presented to users in a unidirectional manner without opportunities for follow-up interactions.
In this work, we explore \emph{interactive} XAI, following growing calls from the AI and HCI communities~\cite{lakkaraju2022rethinking,hohman2019gamut,kulesza2015IUI,Miller2019,zhang2023may,Abdul2018CHI}.

Prior work has demonstrated the needs and benefits of interactive XAI.
Notably, Lakkaraju \etal~\cite{lakkaraju2022rethinking} 
% interviewed practitioners in healthcare and policy and found that they 
found practitioners
strongly prefer interactive interfaces when making decisions with AI systems.
Hohman \etal~\cite{hohman2019gamut} 
% developed a tool for model interpretation and 
found that interactivity was fundamental for data scientists in interpreting and comparing AI systems.
Kulesza \etal~\cite{kulesza2015IUI} 
% built a tool for explanatory debugging and 
found that interactivity increased users' understanding of the AI system and ability to correct its mistakes.

However, there is a lack of interactive XAI tools that help users gain a better understanding of \emph{computer vision} models through direct interaction with the models. 
Many existing tools are proprietary (e.g., AIFinnity~\cite{cabrera2023TOCHI}, Symphony~\cite{Symphony}), not applicable to computer vision models (e.g., Gamut~\cite{hohman2019gamut}, EluciDebug~\cite{kulesza2015IUI}, TalkToModel~\cite{TalkToModel}, AVTALER~\cite{zhou2023interactive}), or support different functionalities (e.g., Shared Interest~\cite{SharedInterest}, ActiVis~\cite{Kahng2017ActiVisVE}, CNN Explainer~\cite{wangCNNExplainerLearning2020}). 
In contrast, our interactive tool, CHM-Corr++, enable users to directly control an image classification model's attention to particular regions of the input and observe changes to its outputs (see \cref{fig:chm_corr++}).
We expect CHM-Corr++ to help users
% With our interactive tool, users can
build an understanding of if, when, and how the model changes its outputs, on top of the understanding provided by static explanations.

Finally, our work builds on and contributes to research on human-AI collaboration~\cite{Lai2022Content,Arous2020opencrowd,Ashktorab2020Game,Cai2019CSCW,Wang2019Datascientist,krishna2022social,Bansal2021Team,bansal2019hcomp,oneill2022teaming,Nguyen2018Factchecking} that explores how humans work together with AI systems to achieve shared goals.
Particularly relevant is work that studied explanations' role in human-AI decision making, especially in the context of fine-grained visual recognition~\cite{kim2022hive,taesiri2022visual,nguyen2021effectiveness,colin2022cannot,Kim2023CHI,nguyen2023advisingnets}.
However, most if not all explanations studied in prior work are \emph{static}.
In this work, we further the field's understanding by exploring the role of \emph{dynamic} explanations in human-AI collaboration.

\section{Method}

% \subsection{Dataset}
% \textbf{CUB-200-2011}~\cite{wah2011caltech} (CUB) represents a task of fine-grained bird-image classification and includes 11,788 images across 200 distinct bird species, with 5,994 for training and 5,794 for testing.

% is a fine-grained, bird-image classification task chosen to complement ImageNet.
% CUB contains 11,788 images (5,994/5,794 for train/test) of 200 bird species.

% \subsection{Classifiers}
\subsection{CHM-Corr++: An interactive interface for controlling model attention}
% \subsection{CHM-Corr++: User-Guided Model for Correspondence-Based Image Classification}
% \subsection{CHM-Corr++: Image Classification Based on User-Guided Attention}
% \subsection{Interactive Tool for Guiding Model Attention}

% In this work, we develop an interactive interface that enables users to control model attention for interactive human-AI collaboration.
For interactive human-AI collaboration, we developed an interactive interface that enables users to control model attention.
Our interface is built on CHM-Corr~\cite{taesiri2022visual},
% \subsec{CHM-Corr generates static explanations}
% The CHM-Corr~\cite{taesiri2022visual} classifier is 
a visual correspondence-based image classification model that produces \textit{static} explanations of its outputs.
% designed to improve image classification accuracy and explainability. 
Given an input image,
% Initially, 
CHM-Corr employs a kNN approach to extract the $N = 50$ most similar candidate images from the training set. 
It then divides the input image 
% These candidates are re-evaluated using a visual correspondence method, which divides the \emph{input image}
into $7\times7$ non-overlapping patches and compares them with corresponding patches in each candidate image.
Based on the cosine similarity among patches, CHM-Corr makes the prediction. %determines the final label.
% CHM-Corr then generates cross-correlation maps to assess the cosine similarity between patches, enabling a comparison that focuses on significant image patches.
% In the final stage, CHM-Corr re-ranks these candidates based on the patch-wise similarity and determines the final image class based on the dominant class among the top-$k$ (where $k = 20$) candidates. 
% CHM-Corr provides \textit{static} explanations for each query and serves as the baseline in our experiments.

However, CHM-Corr is completely automatic and sometimes focuses on image patches that are not semantically meaningful to users (\eg background or indiscriminative features -- Figs.~\ref{fig:static1} \& ~\ref{fig:static3}).
Therefore, we built an interactive interface named CHM-Corr++ that enables users to select image patches that the model should focus on, or in other words, control the model's attention.
At a high level, users are presented with the new attention, support samples, and model outputs. % CHM-Corr classification outputs
The interface was developed using Python and Gradio~\cite{abid2019gradio}. 
We utilized Gradio with a custom HTML component that enables users to control model attention in a $7\times7$ grid. 
% This grid was then be used to find correspondence locations between 
% % each query image and the candidate nearest neighbors.
% the input and candidate images.
% \subsec{CHM-Corr++ generates dynamic, user prompt-based explanations}
% In the original CHM-Corr classifier, significant parts of the image are extracted by calculating the cross-correlation map between the query image and each candidate image. 
% This automated process has been shown to enhance accuracy, but it might not focus on the parts of the image that are most semantically important. 
% Instead of choosing image patches this way, we can ask a human to pick patches manually, and then continue the classification process with the user-provided "attention."
% We call this human prompt-based classifierl CHM-Corr++. \footnote{Not to be confused with CHM-Corr+, a keypoint-based classification model~\cite{taesiri2022visual}}.
Compared with CHM-Corr, CHM-Corr++ enables a more interactive and user-centric image classification process, accompanied by \textit{dynamic} explanations of the model's outputs.

% Both classifiers are based on the reference implementation of CHM-Corr~\cite{taesiri2022visual}. 
% The only modification applied is the expansion of the kNN pool to break ties; when two or more classes among the top-20 images have the same score, the kNN pool is expanded to include more images until the tie is broken.

% \subsection{Problem setup}

% Our goal is to measure the efficacy of both static and dynamic explanations in assisting users to accept or reject the original decision made by AI.
% Let's say the AI model comes with a prediction for an image, and the user's task is to verify whether it is correct or wrong by manually checking the query image, predicted class, and explanations provided by the model.
% CHM-Corr can provide a set of five images from the target predicted class, as well as providing some patch annotations on the image showing the most similar corresponding patch pairs between the query bird and the explanations.
% CHM-Corr++, in addition to showing the same type of explanations, would allow the user to make further predictions by moving the source patches on the query image and seeing how that would change the results.

% \textit{... classify images using static explanation vs dynamic explanation (CHM-Corr vs CHM-Corr++)}

% \subsection{Problem setup}
\subsection{User study}
% We conduct a user study to explore the effectiveness of dynamic vs. static explanations.
We next explore the effectiveness of static and dynamic explanations with a user study.
Our problem setup is as follows (see Fig.~\ref{fig:chm_corr++}): Given an input image (\eg \class{Cardinal}), the model predicts its class {\setlength{\fboxsep}{0pt}\colorbox{gray!15}{\strut(\textbf{c})}} and provides an explanation {\setlength{\fboxsep}{0pt}\colorbox{MyLightGreen}{\strut(\textbf{b})}} for its prediction. The user's task is to accept or reject {\setlength{\fboxsep}{0pt}\colorbox{ResNetBlue}{\strut(\textbf{d})}} the model's original prediction (\class{Summer Tanager}) based on the provided explanation.
% If the model's original prediction is correct, acceptance is a ``correct'' decision and rejection is an ``incorrect'' decision.
% If the model's original prediction is incorrect, rejection is a ``correct'' decision and acceptance is an ``incorrect'' decision.

% Our goal is to measure the efficacy of both static and dynamic explanations in assisting users to accept or reject the original decision made by an AI model. Consider a scenario where an AI model predicts the class of a query image, and the user's task is to verify whether the prediction is correct or incorrect by manually examining the query image, the predicted class, and the explanations provided by the model.

\textbf{Static vs. dynamic explanations.}
In the \emph{static} explanation setting, the model provides five support samples from the predicted class, along with patch annotations highlighting the most similar corresponding patch pairs between the input and the five support samples (see Fig.~\ref{fig:static1}).
% The user is then asked to judge the AI's decision by either accepting or rejecting it based on these static explanations.

In the \emph{dynamic} explanation setting, the model provides the same type of explanations. 
However, users can also control the model's attention by selecting input image patches the model should focus on.
Based on the selections, the model makes a prediction again and produces corresponding explanations (shown in Fig.~\ref{fig:dynamic1}). Note that the new prediction can be same as the original prediction.
% The user can select different image patches, and the model will provide subsequent classifications and explanations based on these modifications.
This process enables users to explore if, when, and how the model's prediction changes based on their selections (\eg users can generate counterfactual explanations via observing the answer for ``what if?'' questions).
% This interactive collaboration between the user and the AI model allows the user to explore how the model's predictions change with different attention configurations and assess the model's sensitivity to these changes.
% After this interaction, the user is asked to accept or reject the original AI prediction based on their understanding gained through the dynamic explanations.
For example, Fig.~\ref{fig:chm_corr++} demonstrates how the model's predictions change upon human interaction.
Initially, using patches that only includes the bird's body, the model (CHM-Corr~\cite{taesiri2022visual}) incorrectly predicts \class{Summer Tanager} (Step 1). However, with user-guided attention on patches that include both the body and the beak, the model correctly predicts \class{Cardinal} (Step 2).
% For example, in Figure~\ref{fig:chm_corr++}, using the patches on the body of the left bird in the image, CHM-Corr++ predicts \class{Summer Tanager} (Step 1). 
% However, if the user modifies the attention to include the patch related to the beak of the bird, the classifier would correctly predict the class as \class{Cardinal} (Step 2).

The key difference between static and dynamic explanations lies in the level of interactivity and user involvement. Static explanations provide a fixed set of supporting information to help users make a decision, while dynamic explanations allow users to actively explore and influence the model's behavior, leading to a more engaging and informative decision-making process.

\textbf{Study materials.}
Following prior works~\cite{nguyen2021effectiveness,taesiri2022visual,colin2022cannot}, we balance the number of correct and incorrect model predictions. 
% We follow previous human studies for XAI in image classification to balance the ratio of correct and incorrect AI predictions~\cite{nguyen2021effectiveness,taesiri2022visual,colin2022cannot}.
From the test set of CUB-200~\cite{wah2011caltech}, we select 600 samples, consisting of 300 correctly classified and 300 misclassified by CHM-Corr~\cite{taesiri2022visual}, resulting in a random-chance accuracy of 50\% for the task.
We input each image to the model to obtain the model's predictions and explanations.
% For each of these images, we then feed to the classifier and extract both the outputs and visual-correspondence explanations.

\textbf{Data collection and participants.}
Due to the complexity of our interface, we decided to pilot the study with 
ML experts who are knowledgeable about XAI.
% people who are knowledgeable about machine learning (ML) and explainable AI (XAI).
We recruited 18 participants, most of whom were Master's and Ph.D. students in ML.
% We reach out to 18 participants, mostly Master's and Ph.D. students, and ask for their help. 
% These volunteers, who are
% They were well-versed in ML concepts such as feature attribution maps and nearest-neighbor algorithms and were familiar with CUB-200. % do the experiments without Internet or help from other people.
% By the end of the study, we end up with 1400 trials, and it is worth noting that a participant could take part in more than one session.
Each participant completed one to several submissions, where each submission consisted of 20 decisions on whether to accept or reject the model's original prediction (\eg \class{Summer Tanager} in Fig.~\ref{fig:chm_corr++}).
In total, we collected data on 1400 decisions.
% Acceptance of a correct model prediction and rejection of an incorrect model prediction is deemed a ``correct'' decision.

% \todo{some random students? a few hundred trials -- a small scale study with 8 participants and total of 615 trails }

% \todo{filter out incomplete users i.e. complete < 20 trials}

\begin{table*}[!hbt]
\centering
\small
% \caption{Per-user mean accuracy (\%) with CHM-Corr and CHM-Corr++ explanations on CUB-200.}
% \caption{Mean and standard deviation of decision accuracy (\%) calculated at the participant level.}
\caption{User study results. We report per-user mean decision accuracy ($\mu$) and standard deviation ($\sigma$) over a study of 18 machine learning experts who generated in total of 70 submissions (each with 20 decisions).}
\label{tab:main_table}
\begin{tabular}{|c|cc|cc|}
\hline
\multirow{2}{*}{Explanation type}      & \multicolumn{2}{c|}{\multirow{2}{*}{Static (CHM-Corr)}}               & \multicolumn{2}{c|}{\multirow{2}{*}{Dynamic (CHM-Corr++)}}              \\
                                  & \multicolumn{2}{c|}{}                  & \multicolumn{2}{c|}{}                  \\ \hline
\multirow{4}{*}{$\mu \pm \sigma$} & \multicolumn{2}{c|}{Overall}                     & \multicolumn{2}{c|}{Overall}                     \\ \cline{2-5}
 & \multicolumn{2}{c|}{72.68 $\pm$ 12.36}                     & \multicolumn{2}{c|}{73.57 $\pm$ 10.42}                     \\ \cline{2-5} 
                                  & \multicolumn{1}{c|}{AI originally correct}           & AI originally incorrect             & \multicolumn{1}{c|}{AI originally correct}           & AI originally incorrect             \\ \cline{2-5} 
                                  & \multicolumn{1}{c|}{85.21 $\pm$ 11.82} & 60.13 $\pm$ 18.66 & \multicolumn{1}{c|}{86.79 $\pm$ 13.16} & 59.39 $\pm$ 15.51 \\ \hline
\# of decisions                            & \multicolumn{1}{c|}{283}               & 277               & \multicolumn{1}{c|}{443}               & 397               \\ \hline
\# of submissions                    & \multicolumn{2}{c|}{28}                                    & \multicolumn{2}{c|}{42}                                    \\ \hline
\end{tabular}
\vspace{-11pt}
\end{table*}
\section{Results}

In our study, participants in the dynamic explanation setting ``used'' (\ie controlling the model's attention and seeing a new prediction) the interactive interface 1.93 times after seeing the model's original prediction. That is, participants in the dynamic explanation setting saw around $3  \approx 1 \text{ (original)} + 1.93 \text{ (new)}$ model predictions on average.
In this section, we explore the effect of this interactivity. 

% \subsection{On bird image classification, interactive XAI is not beneficial for decision making}
\subsection{Interactivity did not improve decision accuracy}

% Old version
% With data collected from the human study, we compute the per-user average accuracy over our controlled, balanced image sets (\ie the ratio of samples where AI correctly classified and misclassified is approx. 50/50) and report in Table~\ref{tab:main_table}.
% Surprisingly, we find that dynamic explanations (CHM-Corr++ users score 73.57\%), which one might naturally assume to be superior, do not show a significant improvement over static ones (CHM-Corr users score 72.68\%) in helping users making more accurate decisions.

% The findings, with a t-statistic of -0.321 and a p-value of 0.749, indicate that the average accuracy levels for users exposed to both types of explanations are not significantly different. 
% This challenges the common belief that dynamic, interactive content inherently boosts user comprehension or performance~\cite{lakkaraju2022rethinking,zhang2023may}. 

% New version
To our surprise, interactivity did not improve participants' decision accuracy. 
Dynamic explanations provided little to no benefit over static explanations for participants in assessing the correctness of the model's original prediction. 
The overall decision accuracy is 72.68\% with static and 73.57\% with dynamic explanations (\cref{tab:main_table}).
Both are higher than the random-chance accuracy (50\%) but still far short of where we want to be (100\%).
This result suggests that interactivity does not always benefit users, contrary to common belief that interactivity inherently boosts user understanding and task performance~\cite{lakkaraju2022rethinking,zhang2023may}.

\subsection{Participants struggled to reject incorrect model predictions}

% Old version
% Table~\ref{tab:main_table} shows that participants identify correct AI classifications way more easily than incorrect ones. 
% Specifically, the mean accuracy for correct classifications with static explanations is 85.21\% with a standard deviation of 11.82, and under the dynamic explanations, it slightly increases to 86.79\% with a standard deviation of 13.16. 
% In contrast, the mean accuracy for identifying wrong classifications is significantly lower, at 60.13\% (±18.66) for static and 59.39\% (±15.51) for dynamic explanations. 
% This difference in accuracy (and large deviations) highlights the challenge in recognizing AI errors compared to correct outcomes as previously highlighted~\cite{fok2023search,nguyen2021effectiveness,taesiri2022visual}, underscoring the necessity for improved interpretability and transparency in AI systems to better detect and understand AI errors~\cite{adebayo2022towards,adebayo2020debugging}.

% New version
Next, to understand where participants struggled most, we separately analyze results on instances where the model's original prediction was correct and instances where it was incorrect (\cref{tab:main_table}).
We find participants' decision accuracy on correct instances is much higher than that on incorrect ones for both types of explanations: 85.21\% vs. 60.13\% with static, 86.79\% vs. 59.39\% with dynamic.
This result is consistent with prior findings that users tend to accept AI predictions as correct even when they are incorrect~\cite{fok2023search,nguyen2021effectiveness,taesiri2022visual,kim2022hive}, highlighting the need for tools that help users detect and reject AI errors~\cite{adebayo2022towards,adebayo2020debugging}.

% \subsection{Verifiability does not emerge with dynamic explanations}

% \begin{figure*}[!ht]
%     \centering
%     \begin{subfigure}[b]{0.49\textwidth}
%         \centering
%         \includegraphics[width=\textwidth]{sec/images/confusion_matrix_CHMCorr.pdf}
%         \caption{}
%         \label{fig:}
%     \end{subfigure}
%     \hfill % This ensures that there is some space between the two subfigures, if needed
%     \begin{subfigure}[b]{0.49\textwidth}
%         \centering
%         \includegraphics[width=\textwidth]{sec/images/confusion_matrix_CHMCorr++.pdf}
%         \caption{}
%         \label{fig:}
%     \end{subfigure}
%     \caption{Confusion matrices for two groups of users using (a) static (CHM-Corr) and (b) dynamic (CHM-Corr++) explanations.}
%     \label{fig:confusion_matrix}
% \end{figure*}

% \subsection{Interactivity could enable verifiability of AI explanations}
% \subsection{Interactivity can help with verification of model predictions}
% \subsection{Interactivity helped participants explore the model's consistency}
% \subsection{Interactivity could improve XAI verifiability}
% \subsection{The usefulness of dynamic explanations depends on the interaction outcomes}
\subsection{The usefulness of interactivity depended on the interaction outcomes}
\label{sec:usefullness}

\begin{table}[b!]
\vspace{-11pt}
\centering
% \caption{CHM-Corr++ user accuracy under different conditions.}
\caption{Participants' decision accuracy (\%) with dynamic explanations under different settings.}
\label{table:accuracy_conditions}
\resizebox{\columnwidth}{!}{
\begin{tabular}{l|l}
\hline
\textbf{AI model correctness \wrt human interaction}  & \textbf{Acc (\%)} \\ \hline
(i)  Originally \textcolor{green!40!black}{{correct}} and consistent (always correct)                                     & 90.80                  \\
(ii) Originally \textcolor{green!40!black}{{correct}} and inconsistent (becomes incorrect)                                   & 75.21                  \\
(iii) Originally \textcolor{red!60!black}{{incorrect}} and consistent (always incorrect)                     & 52.55                  \\
(iv) Originally \textcolor{red!60!black}{{incorrect}} and inconsistent (always incorrect)                       & 62.11                  \\
(v)  Originally \textcolor{red!60!black}{{incorrect}} and inconsistent (becomes correct)                        & 65.43                  \\ \hline
\end{tabular}
}
\end{table}

% New version
Finally, to better understand the effect of interactivity, we break down participants' decision accuracy with dynamic explanations based on the interaction outcomes (\cref{table:accuracy_conditions}).
% To better understand the effect of interactivity, we take a closer look at the results, focusing on XAI verifiability~\cite{fok2023search}. In Table~\ref{table:accuracy_conditions}, we breakdown decision accuracy based on the interaction outcomes to explore if and how dynamic explanations help participants verify the model's predictions. 
Here, ``consistent'' refers to the model maintaining its original prediction even after the user controlled its attention. % of user manipulation of model attention.
% Here, ``consistent'' refers to the model maintaining its original top-1 prediction regardless of user manipulation of model attention, while ``classifiable'' indicates that the model can make a correct prediction when users manipulate the model's attention.

\textbf{When the model is originally correct (i, ii)}, we find that participants' decision accuracy is higher when the model is consistent than not (90.80\% vs. 75.21\%).
This result is in line with our expectations.
When the model maintains its prediction after attention control, participants may gain higher confidence in the prediction and accept it as correct (see Supp. \cref{fig:both_correct,fig:interactivity_hurts}). % \looseness=-1

\textbf{When the model is originally incorrect (iii, iv, v)}, participants' decision accuracy is lower when the model is consistent than not (52.55\% vs. 62.11 $\to$ 65.43\%).
Again, this result is as expected.
When the model maintains its prediction, even when it is incorrect, participants may gain higher confidence in the prediction and accept the prediction as correct.
What happens when the model is inconsistent as shown in Supp. \cref{fig:always_fail}?
We find that when the model's new predictions are always incorrect, participants' decision accuracy is 62.11\%.
But when the model eventually becomes correct, participants' decision accuracy goes up to 65.43\%.
That is, the interactive interface is most helpful when users' attention control changes the model's prediction from incorrect to correct (Supp. \cref{fig:negative_verify}).
As such, understanding when users can and cannot help the model be more accurate, and aiding users in the process, would be important directions for future research.

\section{Discussion}
We assume two leading hypotheses for why dynamic explanations do not surpass static explanations in improving human decision accuracy.
First, regarding the nature of the task, in most instances, AI attention is already sufficient, as the birds are well-centered and clearly visible. 
Changing the task domain, for example, to include complex scenes where AI struggles to focus on the correct pixels, would likely enhance the effectiveness of CHM-Corr++.
Second, we mentioned in Sec.~\ref{sec:usefullness} that CHM-Corr++ is especially helpful when the base CHM-Corr model can classify correctly.
Yet, this base classifier has shortcomings (see Supp.~\ref{sec:bad_chm}) and inherently makes CHM-Corr++ ineffective in many cases (\eg Supp. Figs.~\ref{fig:interactivity_hurts} \& ~\ref{fig:always_fail}).
We hope our open-source tool and investigation of dynamic explanations stimulates further research towards enabling effective human-AI interaction in computer vision.
% Ideally, humans would endeavor to uncover novel knowledge, yet they must also account for the effort required to acquire this information. 
% Therefore, the balance between effort and gain should be carefully weighed as pointed out by Helena \etal.~\cite{vasconcelos2023explanations}.

% verifiability

% model ability

% nature

% \clearpage
\section*{Acknowledgments}
We are grateful for the participation of volunteers who spent their time and efforts in our human studies.
We also thank the anonymous reviewers of XAI4CV workshop for their helpful feedback.
We also thank Travis Thompson, Pooyan Rahmanzadehgervi, and Tin Nguyen from Auburn University for their helpful feedback on our early results.
AN is supported by NaphCare Foundations, Adobe gifts, and NSF grant no. 2145767.

{
    \small
    \bibliographystyle{ieeenat_fullname}
    \bibliography{main}
}

% \clearpage
% \setcounter{page}{1}

% \maketitlesupplementary

\clearpage
\onecolumn % Switch to single column
\setcounter{page}{1}
\begin{center}

\textbf{\large Supplemental Materials}
\end{center}

\newcommand{\beginsupplementary}{%
    \setcounter{table}{0}
    \renewcommand{\thetable}{A\arabic{table}}%
    
    \setcounter{figure}{0}
    \renewcommand{\thefigure}{A\arabic{figure}}%
    
    \setcounter{section}{0}
    % For section headers starting with S
    \renewcommand{\thesection}{A\arabic{section}}
    \renewcommand{\thesubsection}{\thesection.\arabic{subsection}}
}
\beginsupplementary%
\setcounter{figure}{0}
\renewcommand{\thefigure}{A\arabic{figure}}%
\renewcommand{\theHfigure}{SuppFigureA\arabic{figure}} % Unique identifier for figures

\setcounter{table}{0}
\renewcommand{\thetable}{A\arabic{table}}%
\renewcommand{\theHtable}{SuppTableA\arabic{table}} % Unique identifier for tables

\setcounter{section}{0}
\renewcommand{\thesection}{A\arabic{section}}
\renewcommand{\thesubsection}{\thesection.\arabic{subsection}}
\renewcommand{\theHsection}{SuppSectionA\arabic{section}} % Unique identifier for sections

% \clearpage
% \setcounter{page}{1}
% \maketitlesupplementary

\section{The significance t-test for comparing two groups}
\label{sec:t_test}
With data collected from the human study, we compute the per-user average accuracy over our controlled, balanced image sets (\ie the ratio of samples where AI correctly classified and misclassified is approx. 50/50) and report in Table~\ref{tab:main_table}.
Yet, we find that dynamic explanations (CHM-Corr++ users score 73.57\%), which one might naturally assume to be superior, do not show a significant improvement over static ones (CHM-Corr users score 72.68\%) in helping users making more accurate decisions.

The findings, with a t-statistic of -0.321 and a p-value of 0.749, indicate that the average accuracy levels for users exposed to both types of explanations are not significantly different. 
This challenges the common belief that dynamic, interactive content inherently boosts user comprehension or performance~\cite{lakkaraju2022rethinking,zhang2023may}.

\section{The shortcomings of CHM-Corr classifier}
\label{sec:bad_chm}

\begin{figure*}[!hbt]
    \centering
    \includegraphics[width=0.90\linewidth]{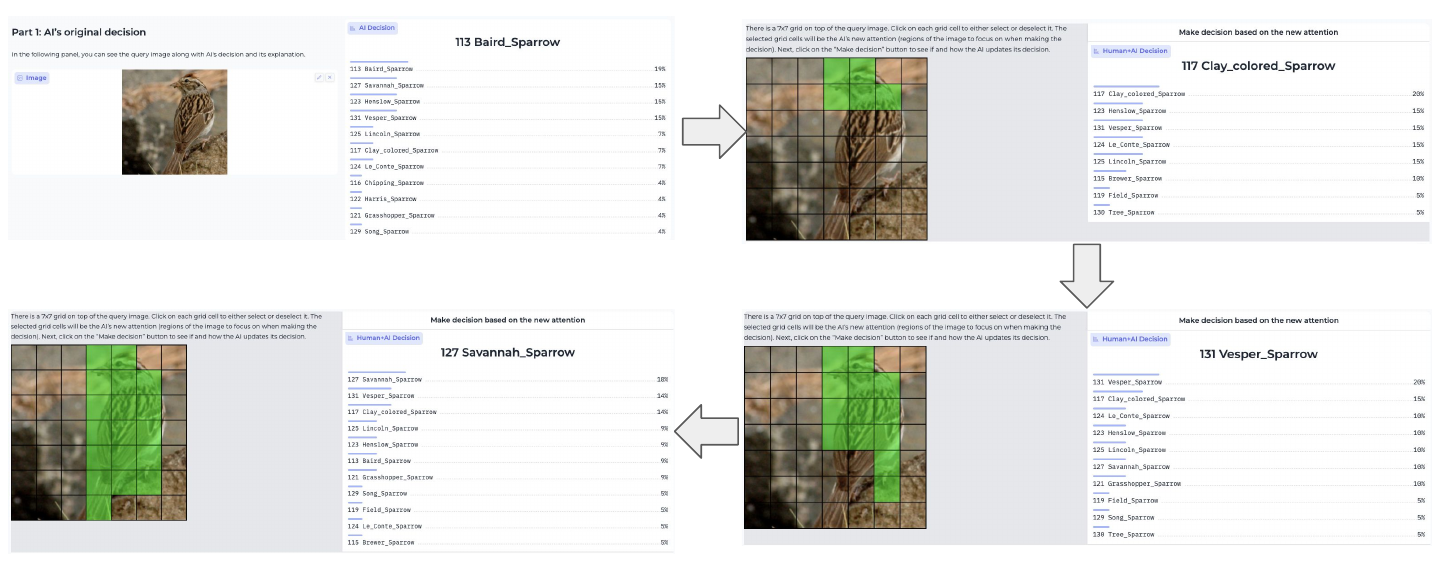}

    \caption{A sample that is unclassifiable for the classifier CHM-Corr. 
    The ground-truth label is \textcolor{green!40!black}{\class{\scriptsize Chipping Sparrow}}.
    Initially, AI makes a wrong classification of \textcolor{red!60!black}{\class{\scriptsize Baird Sparrow}}.
    With user-guided attention, the top-1 label evolves from \textcolor{red!60!black}{\class{\scriptsize Baird Sparrow}} $\to$ \textcolor{red!60!black}{\class{\scriptsize Clay-colored Sparrow}} $\to$ \textcolor{red!60!black}{\class{\scriptsize Vesper Sparrow}} $\to$ \textcolor{red!60!black}{\class{\scriptsize Savannah Sparrow}} but none of them matches the groundtruth, making users unable to make decisions given the explanations.
    }
    \label{fig:always_fail}
\end{figure*}

For some samples, AI fails to classify the input image correctly regardless of how the attention is directed towards the image (see Fig.~\ref{fig:always_fail}).
This indicates that improving attention alone does not suffice for classifiers to accurately classify these samples, suggesting the need for insights into developing new models that focus on more than just improving attention mechanisms. 
Moreover, the underlying nature of the classifier contributes to this issue. 
Given that the classifier employs a k-Nearest Neighbors (kNN) algorithm to retrieve a set of candidate samples, there is a possibility that the ground-truth class may not appear within the candidate pool. 
Consequently, no matter how the CHM-Corr model re-ranks these candidates, it may never correctly identify the top-1 class.

\section{How users interact with explanations}
\label{sec:Qualitative_figures}

\begin{figure*}[!ht]
    \centering
    \begin{subfigure}[b]{0.50\textwidth}
        \centering
        \includegraphics[width=\textwidth]{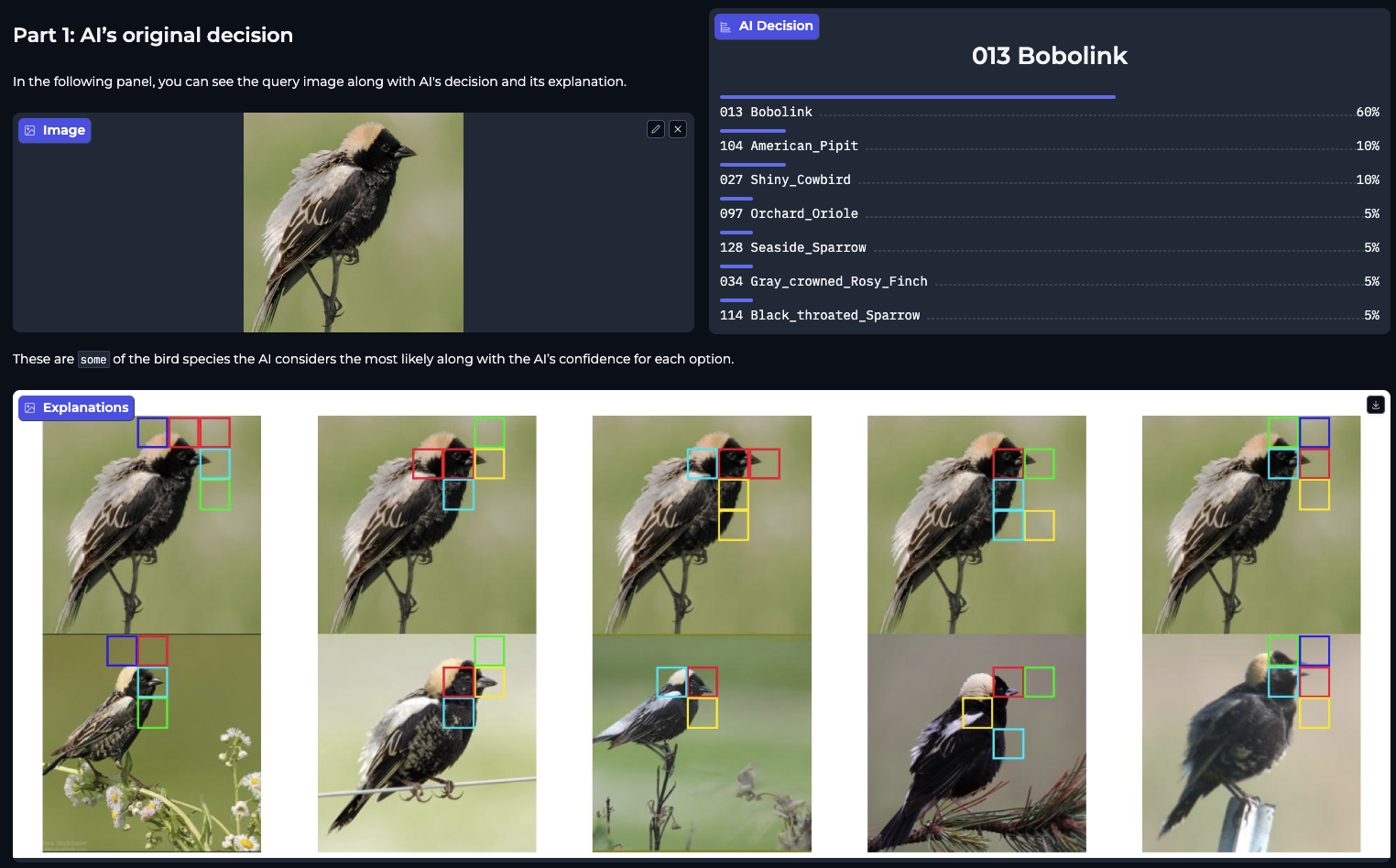}
        \caption{Static explanation for \textcolor{green!40!black}{\class{\scriptsize Bobolink}}.}
        \label{fig:static1}
    \end{subfigure}
    \hfill % This ensures that there is some space between the two subfigures, if needed
    \begin{subfigure}[b]{0.48\textwidth}
        \centering
        \includegraphics[width=\textwidth]{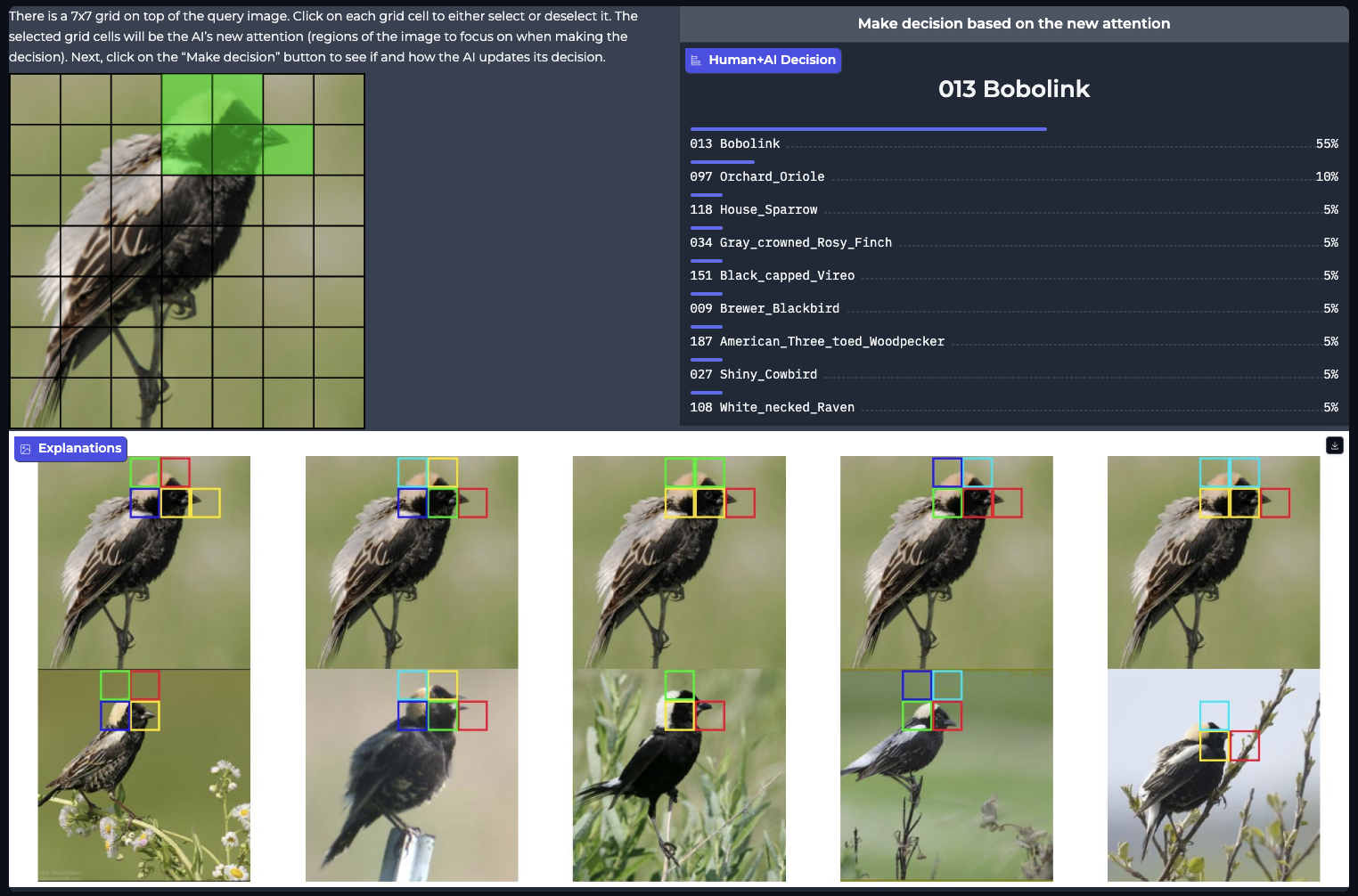}
        \caption{Human-prompt, correspondence-based explanation for \textcolor{green!40!black}{\class{\scriptsize Bobolink}}.}
        \label{fig:dynamic1}
    \end{subfigure}
    \caption{Both dynamic and static explanations enable human users to verify that the AI is predicting the top-1 label correctly.}
    \label{fig:both_correct}
\end{figure*}

\begin{figure*}[!ht]
    \centering
    \begin{subfigure}[b]{0.50\textwidth}
        \centering
        \includegraphics[width=\textwidth]{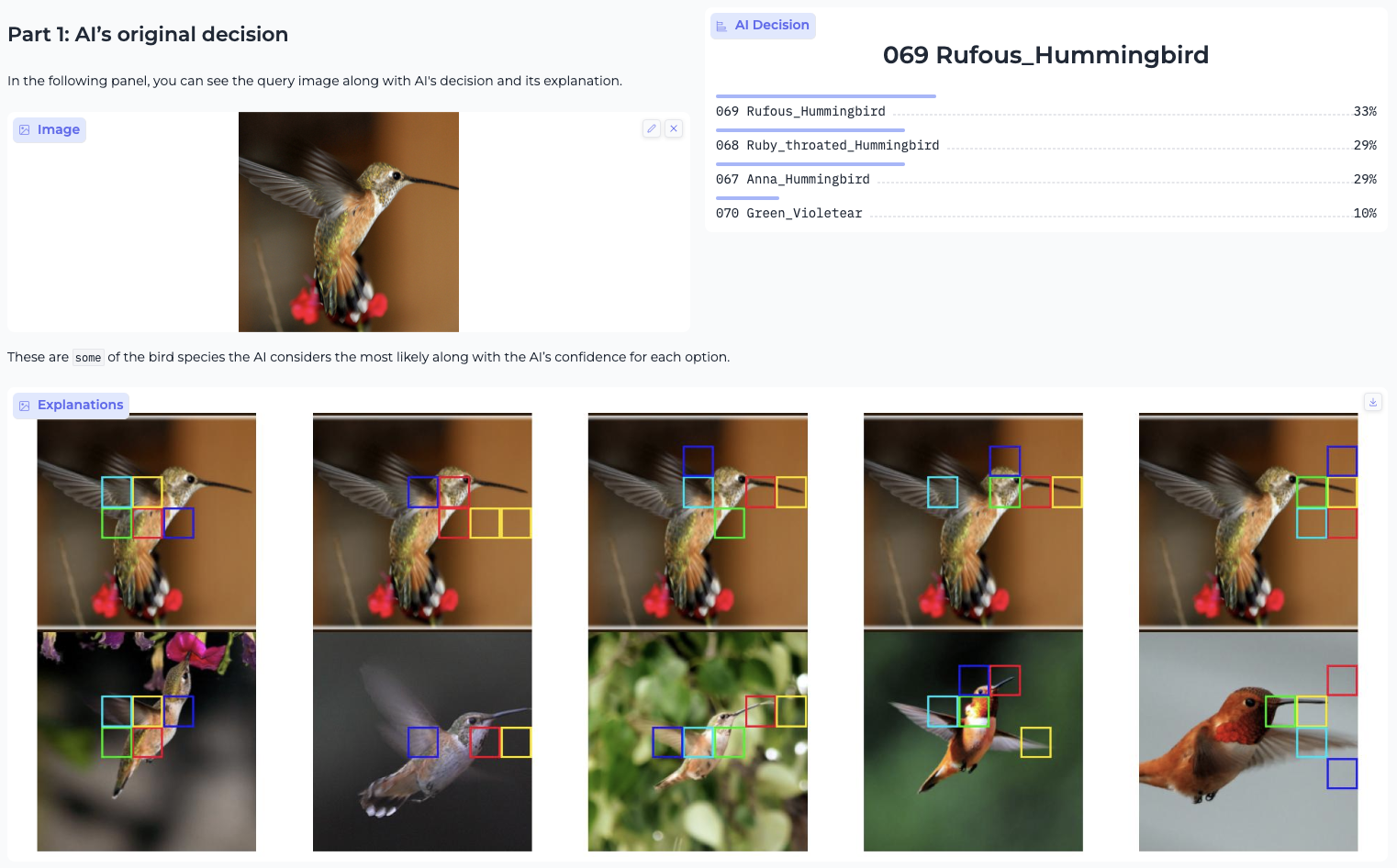}
        \caption{Static explanation for \textcolor{green!40!black}{\class{\scriptsize Rufous Hummingbird}}.}
        \label{fig:static2}
    \end{subfigure}
    \hfill % This ensures that there is some space between the two subfigures, if needed
    \begin{subfigure}[b]{0.48\textwidth}
        \centering
        \includegraphics[width=\textwidth]{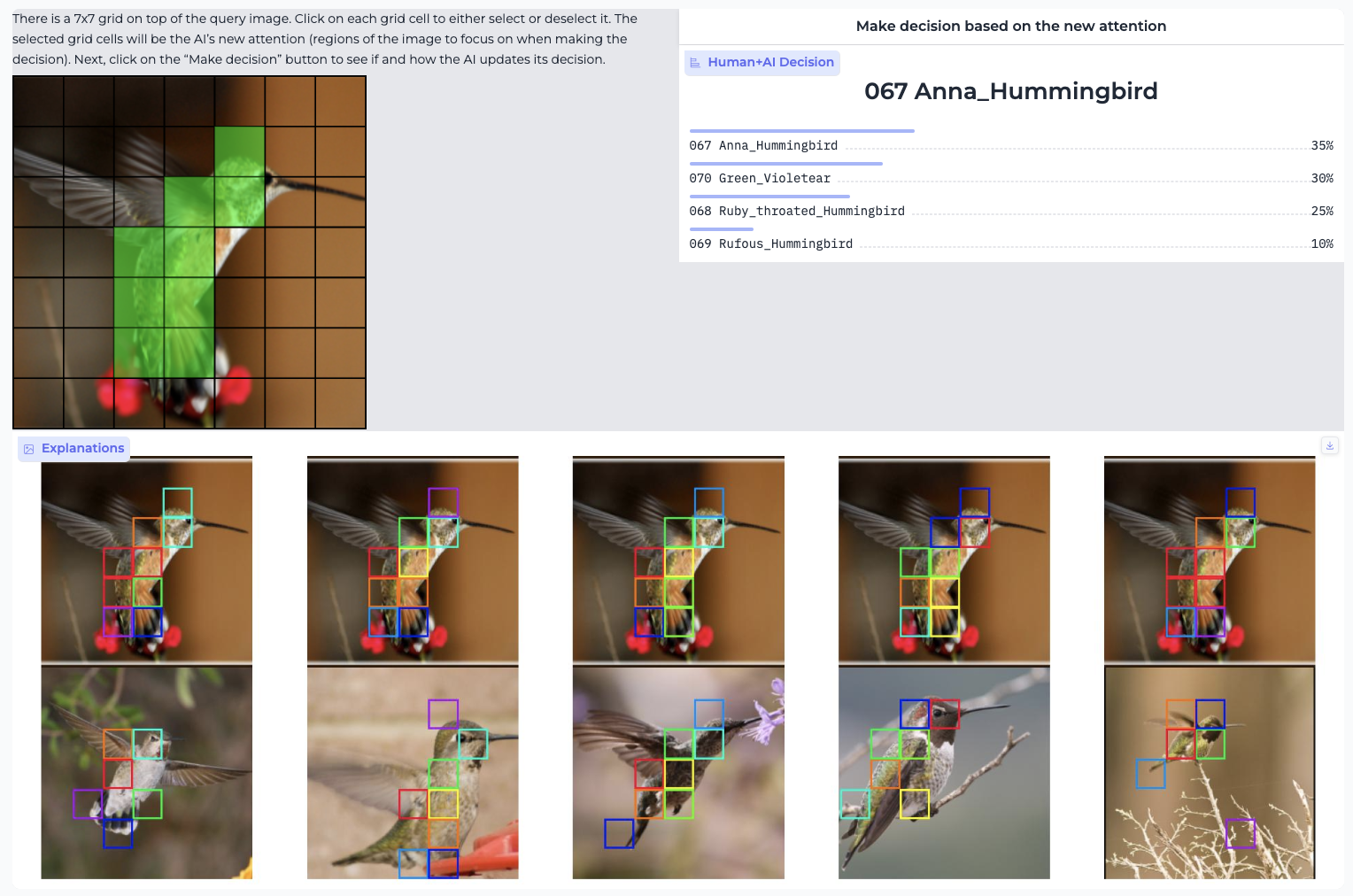}
        \caption{Human-prompt, corr-based explanation for \textcolor{red!60!black}{\class{\scriptsize Anna Hummingbird}}.}
        \label{fig:dynamic2}
    \end{subfigure}
    \caption{Human intervention changes the top-1 label from \textcolor{green!40!black}{\class{\scriptsize Rufous Hummingbird}} $\to$ \textcolor{red!60!black}{\class{\scriptsize Anna Hummingbird}} that makes users more likely to reject the original, correct label.}
    \label{fig:interactivity_hurts}
\end{figure*}

\begin{figure*}[!ht]
    \centering
    \begin{subfigure}[b]{0.50\textwidth}
        \centering
        \includegraphics[width=\textwidth]{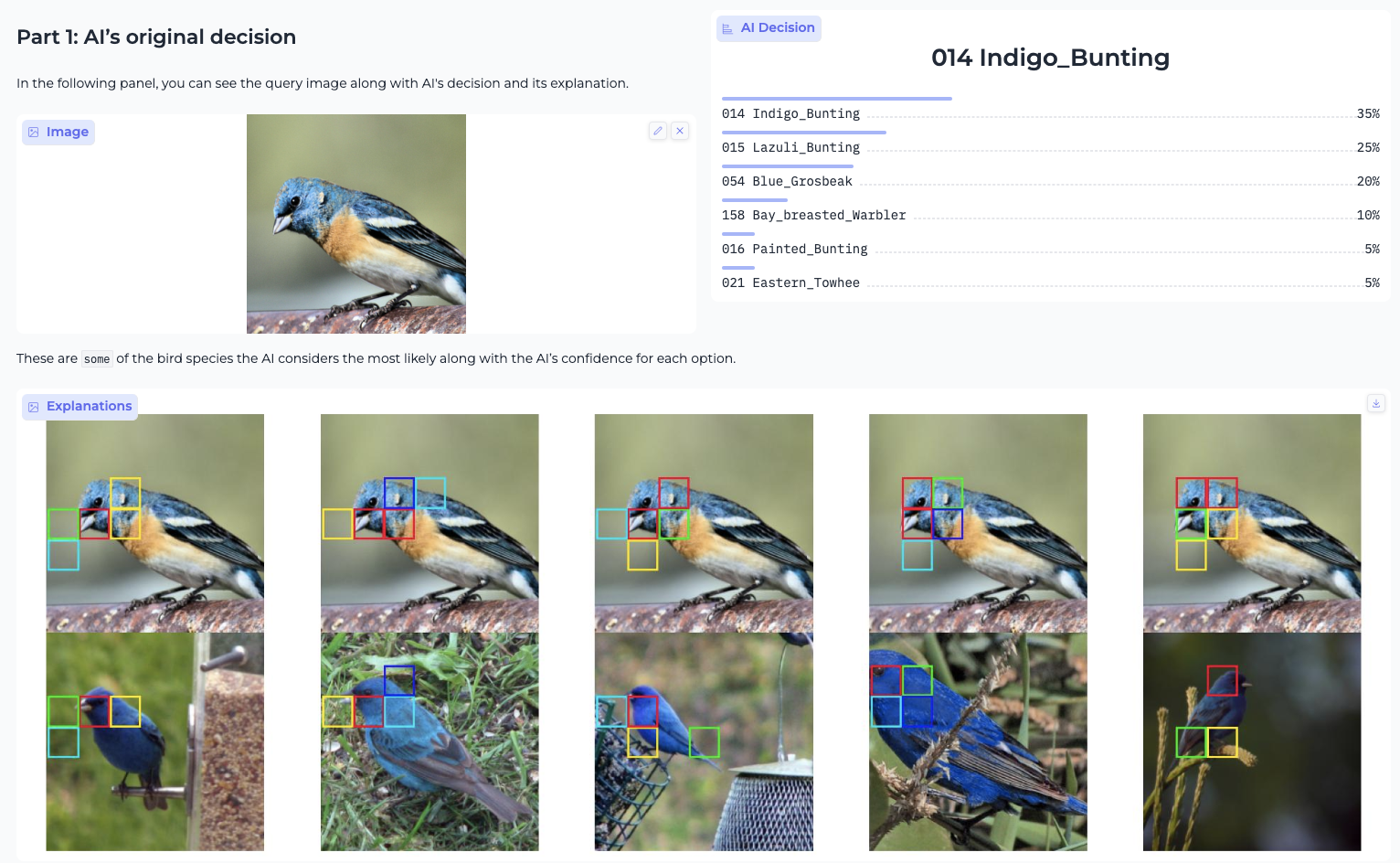}
        \caption{Static explanation for \textcolor{red!60!black}{\class{\scriptsize Indigo Bunting}}.}
        \label{fig:static3}
    \end{subfigure}
    \hfill % This ensures that there is some space between the two subfigures, if needed
    \begin{subfigure}[b]{0.48\textwidth}
        \centering
        \includegraphics[width=\textwidth]{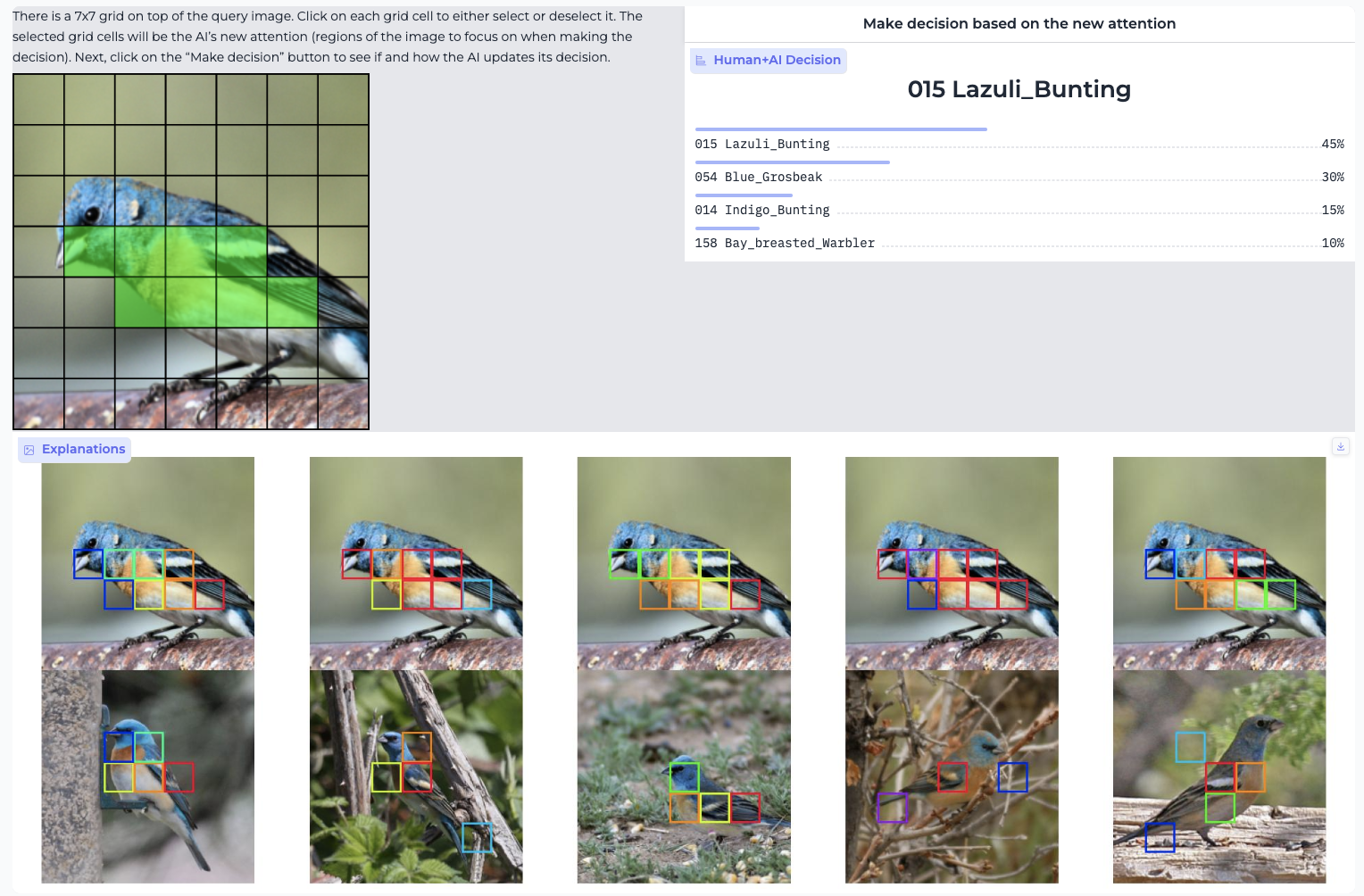}
        \caption{Human-prompt, correspondence-based explanation for \textcolor{green!40!black}{\class{\scriptsize Lazuli Bunting}}.}
        \label{fig:dynamic3}
    \end{subfigure}
    \caption{AI initially makes the wrong classification \textcolor{red!60!black}{\class{\scriptsize Indigo Bunting}} on the input image.
    Human intervention changes the top-1 label from \textcolor{red!60!black}{\class{\scriptsize Indigo Bunting}} $\to$ \textcolor{green!40!black}{\class{\scriptsize Lazuli Bunting}}, a more similar-looking class, encouraging users to reject the original, predicted label.}
    \label{fig:negative_verify}
\end{figure*}

\end{document}